\pdfoutput=1

\documentclass[11pt]{article}

\usepackage{ACL2023}

\usepackage{times}
\usepackage{latexsym}

\usepackage[T1]{fontenc}

\usepackage[utf8]{inputenc}

\usepackage{microtype}

\usepackage{inconsolata}
\usepackage{bm}
\usepackage{amsmath}
\usepackage{amssymb}
\usepackage{multirow}

\usepackage{graphicx}
\usepackage{subfigure}

\title{CIF-PT: Bridging Speech and Text Representations for Spoken Language Understanding via Continuous Integrate-and-Fire Pre-Training}

\author{Linhao Dong\thanks{\ \ Equal contribution.},
 ~Zhecheng An\footnotemark[1], 
 ~Peihao Wu, 
 ~Jun Zhang,
 ~Lu Lu,
 ~Zejun Ma\\
ByteDance AI Lab \\
\texttt{\{donglinhao, anzhecheng, wupeihao, zhangjun.jarry,} \\
\texttt{lulu.0314, mazejun\}@bytedance.com} 
}

\begin{document}
\maketitle
\begin{abstract}
Speech or text representation generated by pre-trained models contains modal-specific information that could be combined for benefiting spoken language understanding (SLU) tasks. In this work, we propose a novel pre-training paradigm termed Continuous Integrate-and-Fire Pre-Training (CIF-PT). It relies on a simple but effective frame-to-token alignment: continuous integrate-and-fire (CIF) to bridge the representations between speech and text. It jointly performs speech-to-text training and language model distillation through CIF as the pre-training (PT). Evaluated on SLU benchmark SLURP dataset, CIF-PT outperforms the state-of-the-art model by 1.94\% of accuracy and 2.71\% of SLU-F1 on the tasks of intent classification and slot filling, respectively. We also observe the cross-modal representation extracted by CIF-PT obtains better performance than other neural interfaces for the tasks of SLU, including the dominant speech representation learned from self-supervised pre-training.
 
\end{abstract}

\section{Introduction}
Spoken language understanding (SLU) plays a key role in speech interaction systems
such as spoken dialogue systems, voice assistants, automated calling robots, etc. 
It focuses on extracting key information and making predictions
from audio signals of human speech \cite{wang2005spoken,Tur2011Spoken}. Traditional methods decompose SLU into two 
cascading tasks: automated speech recognition (ASR) and natural language understanding (NLU),
where audio signals are first transcribed into texts, 
and then processed by a text-based language understanding model.
In the cascading scheme,
the errors of ASR module will be accumulated in the NLU module and degrade the final performance.
Moreover, predicted text of ASR module may not be the ideal interface
for the language understanding task. 
For example, acoustic information
such as intonation and pitch that may be helpful for understanding tasks are lost after ASR.
To tackle the problems above, resent researches employ end-to-end approaches for SLU 
\cite{Serdyuk2018Towards,haghani2018audio,chung2021splat,arora2022espnet}, 
where the language understanding is directly performed from audio signals without explicitly utilizing predicted text of ASR. 

For text-based language understanding tasks, pre-trained language models 
such as BERT \cite{devlin-etal-2019-bert}, RoBERTa \cite{Liu2019-roberta} and GPT \cite{radford2019gpt}
have achieved remarkable success.
These models utilize self-supervised pre-training on large-scale unlabeled corpora to learn
contextual representations in token or sentence level 
with rich syntactic and semantic knowledge \cite{Liu2019linguistic}, 
which significantly benefit downstream tasks such as NLU during fine-tuning.
This self-supervised pre-training fashion has been extended into the representative learning on speech.
Researches such as wav2vec \cite{baevski2020wav2vec}, HuBERT \cite{hsu2021hubert} and data2vec \cite{Baevski2022}
focus on learning better frame-level contextual representations using unlabeled speech data, to improve
the performance of ASR as well as other speech processing tasks. For end-to-end SLU, 
these self-supervised speech models have been proven to be powerful backbones on learning semantic representations
\cite{wang2021fine,arora2022espnet}. 

The self-supervised pre-training methods for speech mainly focus on 
leveraging speech data to model acoustic information \cite{chung2021splat} on the frame level, 
while pre-trained language models work on higher token or sentence levels 
to encode linguistic knowledge \cite{Liu2019linguistic}. 
These two kinds of representation could be combined for better benefiting downstream tasks such as SLU.
The combination of speech and text representations can be performed by jointly pre-training on data of
the two modalites \cite{Chuang2020speechbert}, 
or distillating one pre-trained representations into another \cite{Kim2021two-stage}.
In either way the frame-level speech representation needs to be aligned with the token-level textual representation.
Frame-to-token alignment methods such as forced alignment has been applied 
to speech-text joint pre-training \cite{Chuang2020speechbert}. 
However, these alignment methods mainly rely on external models or rules, 
and can only generate hard alignment mapping that can not be updated in end-to-end training. 
On the other hand, aligning frames and tokens through cross-attention \cite{arora2022espnet,zhu2022cross} 
suffers from high complexity 
and lack of token timestamps that synchronized to frames.

The frame-to-token alignment also plays a critical role in ASR systems.
Various works, such as Connectionist Temporal Classification (CTC) \cite{graves2006ctc}, Listen, Attend and Spell (LAS) \cite{Chan2016las}, 
RNN Transducer (RNN-T) \cite{graves2012rnnt} and 
Continuous Integrate-and-Fire (CIF) \cite{dong2020cif}, 
focus on bringing effective alignment methods for better speech recognition performance.
Among these works, the CIF alignment, which explicitly aggregates frame-level
speech representations into token-level, is adopted in our work to combine with text representation. 
Specifically, we propose a novel pre-training paradigm: Continuous Integrate-and-Fire Pre-Training (CIF-PT) for end-to-end SLU. 
Two pre-training tasks are included in CIF-PT: the first task is speech-to-text modeling \citep{wang2020fairseq} with CIF alignment. In this work, ASR task that transcribes speech to text is applied. The second task is language model distillation (LMD). Since the integrated speech representation by CIF is at token-level, token-level distillation from a pre-trained language model can be performed to inject text-based linguistic knowledge into the representation.
Through the joint pre-training of the two tasks, 
CIF-PT is able to generate representations with information from both speech and text modalites.

We examine our CIF-PT methods in downstream SLU tasks including intent classification and slot filling. 
On SLU benchmark SLURP \cite{bastianelli2020slurp} dataset, 
the end-to-end SLU model with CIF-PT outperforms the state-of-the-art model by 1.94\% of accuracy and 2.71\% of SLU-F1 on the tasks of intent classification and slot filling, respectively.
The cross-modal representation extracted by CIF-PT also shows its competitiveness in comparison of other neural interfaces \citep{rao2020speech, Raju2022} utilized in SLU. The obtained results and a series of experiments including ablation study and the pre-training on out-of-domain data demonstrate the effectiveness and generalization of CIF-PT. 

\section{Related Works}
\paragraph{End-to-End SLU}
Various works extend models originally designed for ASR into the field of SLU.
\citet{peng2022branchformer} propose Branchformer as an alternative to Conformer \cite{Gulati2020}, 
and show performance gains in SLU as well as ASR. 
\citet{huang2022mtl} jointly train ASR and SLU as multitasks to exploit shared knowledge from different tasks.
\citet{seo2022integration} use the probability distribution output of ASR model as continuous token interface (CTI) 
for downstream NLU.
Self-supervised representative learning on speech data provides powerful backbones such as 
wav2vec 2.0 \cite{baevski2020wav2vec}, HuBERT \cite{hsu2021hubert} for SLU. 
\citet{arora2022espnet} propose ESPnet-SLU and
analyze the performance of HuBERT encoder pre-trained with ASR as feature extractor for SLU. 
\citet{wang2021fine} perform partial fine-tuning and entire fine-tuning on pre-trained wav2vec 2.0
and HuBERT on SLU tasks. 

\paragraph{Cross-Modal Pre-training for SLU}
In order to exploit information from speech and text for SLU,
jointly pre-training on both of speech and text data has been proposed. 
SpeechBERT \cite{Chuang2020speechbert} extends the masked language model (MLM) pre-training from BERT
into the mixture of audio and text data. 
In SPLAT \cite{chung2021splat}, a speech module and a language module are jointly pre-trained with
token-level and sentence-level alignment. 
Another branch of researches focus on knowledge distillation from pre-trained language model into pre-trained
speech encoder. 
\citet{Kim2021two-stage} utilize BERT as a teacher to perform sentence-level knowledge distillation 
at the pre-training stage and target-specific distillation during fine-tuning. 
\citet{zhu2022cross} introduce cross-attention between text and speech and perform distillation on the 
attention heads for knowledge transfering. 

\paragraph{Frame-to-Token Alignment in SLU}
In SpeechBERT \cite{Chuang2020speechbert}, forced alignment based on external ASR engine is used to 
train the initial phonetic-semantic joint embedding. 
\citet{chung2021splat} adopt a heuristic alignment approach in SPLAT, where alignment scores
is computed by the cosine similarity between the output embeddings of the pre-trained speech and text models.
The cross-attention alignment is introduced in \cite{zhu2022cross} to capture the interactions
between text tokens and speech frames. For SpeechT5, since the pre-training does not strictly rely on
audio-text pair data, \cite{ao2022speecht5} adopt shared codebook for speech and text representation
and a diversity loss to encourage the alignment in latent space. 

\section{Method}


\begin{figure*}[] 
	\centering 
	\includegraphics[width=\textwidth]{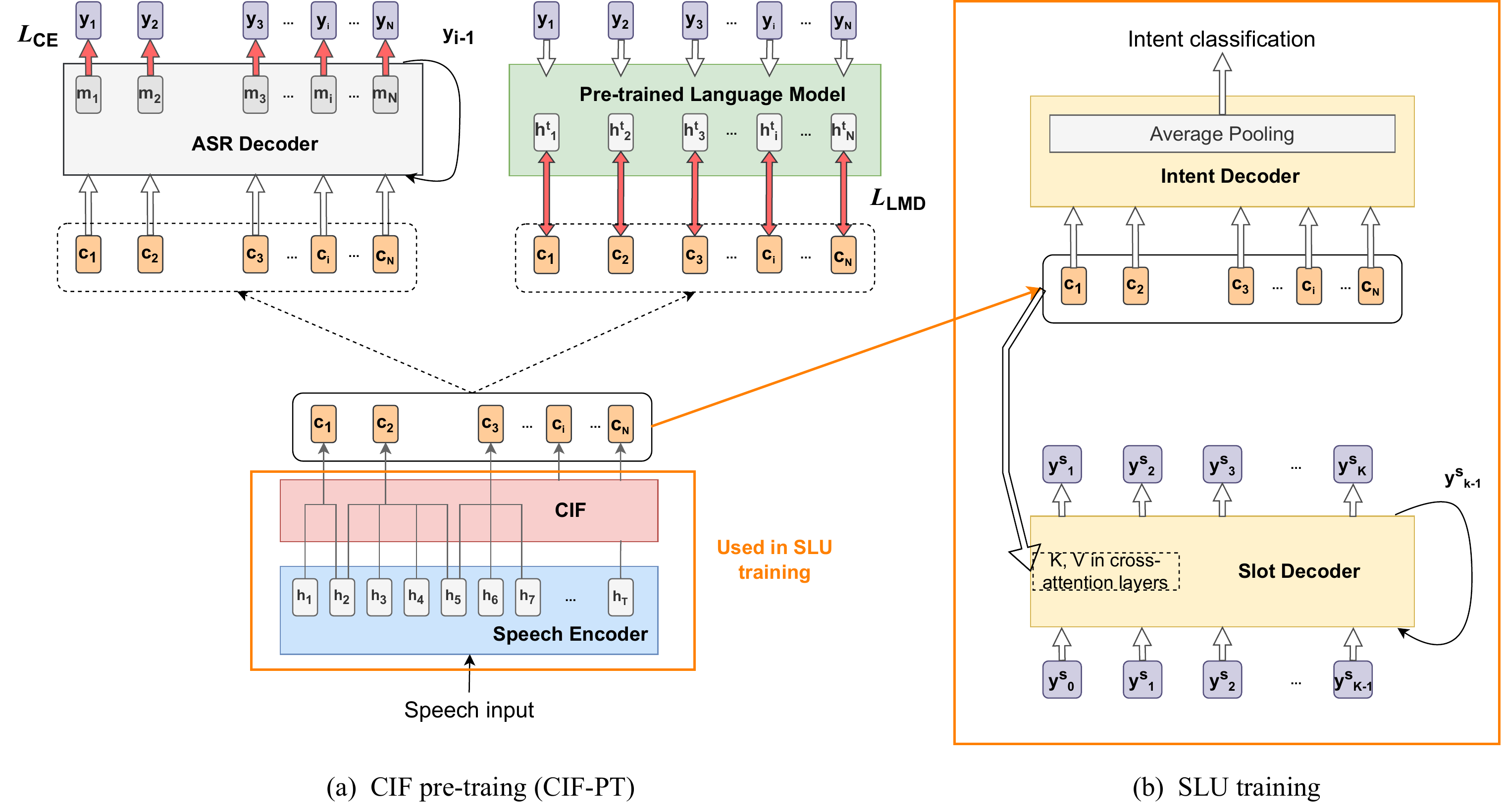}
	\caption{Architecture of our end-to-end SLU model with CIF-PT: (a) shows the procedure of CIF-PT including the ASR task with CIF alignment and token-level language model distillation; (b) shows the model structure of SLU decoder used for SLU training, including intent decoder and slot decoder for IC and SF, respectively.}
	\label{fig_framework} 
\end{figure*}

In this section, we present the architecture of our proposed continuous integrate-and-fire pre-training (CIF-PT) method for SLU. 
As shown in Figure \ref{fig_framework}, our end-to-end SLU models go through two stages: CIF-PT and SLU training. 

During CIF-PT, we employ two pre-training tasks:
ASR training with CIF alignment and token-level language model distillation (LMD). 
These two tasks help the model learn contextual representation of the speech features aligned to the tokens with high level linguistic knowledge. 
After CIF-PT, the pre-trained parameters including the speech encoder and CIF part are used for downstream SLU tasks such as intent classification and slot filling. 

\subsection{ASR training with CIF Alignment}
As shown in Figure \ref{fig_framework}(a), the structure of CIF-based ASR model includes three parts: speech encoder, CIF part, and the corresponding decoder. 
For an input speech utterance, it is first processed into a sequence of frames 
$\bm{x}=[x_1, x_2, \cdots, x_{T^{'}}]$ with length $T^{'}$ via speech feature extractor (e.g. mel-filter bank, convolutional front-end \citep{baevski2020wav2vec}), where $x_t$ is the feature vector of the $t$ th frame.
The speech encoder converts the frame-level input vector into frame-level hidden states:
\begin{equation*}
\bm{h}=[h_1, h_2, \cdots, h_T] = \texttt{enc}([x_1, x_2, \cdots, x_{T^{'}}])
\end{equation*}


CIF part follows the speech encoder to convert the frame-level hidden states $\bm{h}$ into token-level speech representations $\bm{c}$. 
We follow the CIF setup from \citet{dong2020cif}, which is briefed as follows. At first, the encoded hidden states 
$\bm{h}=[h_1, h_2, \cdots, h_T]$ 
are fed into a weight estimator module to calculate a series of weights 
$\bm{\alpha}=[\alpha_1, \alpha_2, \cdots, \alpha_T]$.
The weights $\bm{\alpha}$ and the frame-level hidden states $\bm{h}$ are input to CIF to obtain $\bm{c}=[c_1, c_2, \cdots, c_i, \cdots, c_N]$, where $N$ is the number of total tokens.
Each token-level representation $c_i$ is a linear combination of frame-level representations $\{h_t\}$.
At each frame step $t$, the weight $\alpha_t$ added to an accumulated weight 
$\alpha^a_i\leftarrow\alpha^a_i + \alpha_t$, and 
the frame-level hidden state $h_t$ is integrated into token-level representation $c_i\leftarrow c_i + \alpha_t h_t$,
until the accumulated weight $\alpha^a_i$ exceeds a threshold $\beta$. 
When $\alpha^a_i$ exceeds $\beta$, the weight of the boundary hidden state is divided into two parts $\alpha_t=\alpha_{t1} + \alpha_{t2}$, 
to ensure the accumulated weight for each token is exactly $\beta$, 
and the second part $\alpha_{t2}$ is accumulated to the next token representation.
In such way, the frame-level hidden states are integrated into token-level representation, which not only reduces the redundancy of speech information but also reduces computation complexity when used for the subsequent ASR decoder and downstream understanding tasks.  

We use the autoregressive ASR decoder in \citep{dong2020cif}.
It accepts previous token $y_{i-1}$ and the integrated $c_i$ from CIF part as inputs, and autoregressively predicts the token output distribution for each $c_i$. The CIF-based encoder-decoder model is trained with a cross entropy (CE) loss in a teacher-forcing manner:
\begin{equation*}
\mathcal{L}_{\text{CE}}=\sum_{i=1}^N \log p(y_i|y_{<i}, c_i).
\end{equation*}
Optionally, $\mathcal{L}_{\text{CTC}}$ can be applied on the frame-level hidden states $\bm{h}$ to be jointly trained.
The quantity loss $\mathcal{L}_{\text{QUA}}$ is to supervise the CIF part to predict the quantity of tokens closer the number of target tokens:
\begin{equation*}
\mathcal{L}_{\text{QUA}}=\bigg|\sum_{i=1}^T \alpha_i - N\bigg|.
\end{equation*}
The final CIF loss is the weighted sum of three:
\begin{equation}
\mathcal{L}_{\text{CIF}}=\mathcal{L}_{\text{CE}} + 
\lambda_1\mathcal{L}_{\text{CTC}} + \lambda_2\mathcal{L}_{\text{QUA}}.
\end{equation}

\subsection{Language Model Distillation}
Since the speech representation $c_i$ integrates speech information into the token-level, we use a pre-trained BERT model as a knowledge distillation teacher to inject textual knowledge into speech representation.
Let $\bm{x}=\{x_i\}_{i=1}^T$ be the speech frame sequence 
and $\bm{y}=\{y_i\}_{i=1}^{N}$ be the corresponding transcript token sequence.
As shown in Figure \ref{fig_framework}, 
$\bm{x}$ is encoded into speech feature $\{c_i\}_{i=1}^N$ by the speech encoder and the CIF part. 
$\bm{y}$ is encoded by BERT into contextual representation vectors $\{h_i^{t}\}_{i=1}^N$.
Since $\{c_i\}$ are aligned to tokens, directly token-level knowledge distillation can be performed 
to make the speech representation close to the contextual representation brought by BERT, thus forming a cross-modal representation.

We consider three types of language model distillation (LMD) loss in our paper, MSE loss, smoothed L1 loss and contrastive loss.
Using BERT hidden output $h_i^t$ as target, the MSE loss of $c_i$ is
$\mathcal{L}_{\text{LMD}}^{\text{MSE}}(h_i^t, c_i)=\|h_i^t - c_i \|^2$.
The smoothed L1 loss is proposed in \cite{Baevski2022}, where a $\gamma$ is used to control the transition
from a squared loss to an $L_1$ loss, i.e.
\begin{equation*}
\mathcal{L}_{\text{LMD}}^{\text{SL1}}(h_i^t, c_i)=
\begin{cases}
\frac{1}{2}(h_i^t - c_i)^2 / \gamma & |h_i^t-c_i| \leq \gamma \\
(|h_i^t-c_i| - \frac{1}{2}\gamma) & \text{otherwise.}
\end{cases} 
\end{equation*}
The contrastive loss encourage $c_i$ to be closer to $h_i^t$ than other $c'$ sampled from an in-batch negative
set $\mathcal{N}_c$.
\begin{equation*}
\mathcal{L}_{\text{LMD}}^{\text{cont}}(h_i^t, c_i) = \frac{\exp [\text{sim}(h_i^t, c_i)/\tau]}{\sum_{c'\in \mathcal{N}_c}\exp[\text{sim}(h_i^t, c')/\tau]},
\end{equation*}
where $\text{sim}(\cdot, \cdot)$ is the cosine similarity function and $\tau$ is the temperature scalar.

The LMD task is trained 
simultaneously with CIF-based ASR training as multitasks, which forms the training loss $\mathcal{L}$ of CIF-PT as follows:
\begin{equation}
\mathcal{L}=\mathcal{L}_{\text{CIF}}+\lambda \mathcal{L}_{\text{MLD}}.
\end{equation}

\subsection{Spoken Language Understanding}
After CIF-PT, the pre-trained speech encoder and CIF part convert speech input into the sequence of cross-modal representation $\{c_i\}$, which is used for downstream SLU training.
We evaluate our pre-trained model on SLU tasks of intent classification and slot filling. The corresponding intent decoder and slot decider are shown in Figure \ref{fig_framework}(b).

For intent classification, $\{c_i\}_{i=0}^{N}$ is fed into additional Transformer layers to
generate task specific decoder states. We use the average of decoder state on all position as the utterance representation for intent prediction through a linear projection. 

The slot filling task is performed in a sequence generation style. 
The slot types and slot values are concatenated as targets $\{y^s_i\}$ to train a sequence-to-sequence model, i.e.
``\texttt{[SEP]} slot\_type1 slot\_value1 \texttt{[SEP]} slot\_type2  slot\_value2''.
The slot decoder consists of Transformer decoder layers where the sequence of $c_i$ is used as the key and value of the cross-attention layer.
We train the encoder-decoder to generate slot target sequence $\{y^s_i\}_{i=0}^{K}$
with teacher-forcing.

\section{Experimental Setup}
\subsection{Dataset and Preprocessing}
We conduct experiments on the dataset of SLURP \citep{bastianelli2020slurp}, which is currently the largest SLU benchmark and is also linguistically more diverse than other datasets. It is collected for developing an in-home personal robot assistant. The train, development and test sets split in the SLURP paper are used for the training and evaluation of our methods. In addition to use the in-domain SLURP data for pre-training, we also introduce the Librispeech \citep{panayotov2015librispeech} dataset that contains 960 hours of speech derived from audiobooks as the out-domain pre-training dataset, which is only used in Section \ref{ood_data}. All speech data is re-sampled or kept at 16 kHz, and all text data is converted into a sequence of subword units by the subword-nmt \citep{sennrich2016neural} toolkit \footnote{https://github.com/rsennrich/subword-nmt}. Specifically, we generate 10706 subword units by performing 36000 merge operations on the training set of Librispeech datasets, and use the learned BPE as the only tokenizer for text of all datasets.

\subsection{Model Configuration}
In this part, we detail the model structure and configuration utilized in our experiments. All the models are implemented using \citep{paszke2019pytorch}: 

\paragraph{Encoder} we use two types of speech encoder which are denoted as conformer and data2vec in subsequent experiments. For the encoder of conformer, it consists of a two-layer convolutional front-end and 15-layer conformer blocks \citep{li2021better}. It applies a 8-time temporal down-sampling similar to \citep{dong2019self}. 
The hidden size in the conformer block uses 400. For the encoder of data2vec, it follows the official data2vec-large configuration \citep{baevski2022data2vec} and uses the released model \footnote{https://huggingface.co/facebook/data2vec-audio-large} from \citep{wolf2020transformers}. 
For the text encoder that provides text representation in CIF-PT, we follow the BASE configuration of BERT \citep{devlin-etal-2019-bert} and use our learned BPE tokenizer to perform pre-training on the English Wikipedia corpus.

\paragraph{CIF part} we follow the implementation of weight estimator and CIF calculator in \citep{dong2020cif}. 
The channel number in convolutional layer keeps the same as the hidden size in decoder. The threshold $\beta$ during CIF calculation is set to 1.0. The corresponding scaling strategy and tail handling methods are also used.

\paragraph{Decoder} we use three types of decoder in our experiments, including the ASR decoder for speech-to-text training in CIF-PT, the down-streaming intent decoder for IC and slot decoder for SF. For ASR decoder, it uses the original autoregressive decoder \citep{dong2020cif} with  2-layer self-attention networks (SANs, also known as transformer encoder layers \citep{vaswani2017attention}). The hidden size is 400 when the encoder uses conformer and 512 for data2vec. For intent decoder, it uses 2-layer SANs and a following average pooling layer 
. For slot decoder, it uses 4-layer SANs for the tag-based slot decoder and uses 4-layer transformer decoder layers (with additional cross-attention layer) for the generation-based slot decoder. Without specific statement, the generation-based slot decoder is used by default.
The hidden size keeps the same as ASR decoder for the two types of SLU decoder.

\begin{table}
\centering
\resizebox{0.5\textwidth}{!}{
\begin{tabular}{lcc}
\hline
\textbf{} & \textbf{IC} & \textbf{SF}  \\
\textbf{} & (Acc.) & (SLU-F1)  \\
\hline

MTL-SLT \citep{huang2022mtl} & 83.10\% & 74.49\% \\ 
Speech-Brain \citep{ravanelli2021speechbrain} & 85.34\% & 74.26\% \\
ESPNET-SLU \citep{arora2022espnet} & 86.30\% & 71.90\% \\ 
CTI \citep{seo2022integration} & 86.92\% & 74.66\% \\
Branchformer \citep{peng2022branchformer} & 88.10\% & 77.70\% \\ 
Hubert SLU \citep{wang2021fine} & 89.38\% & 78.92\% \\ 
\hline \hline 
CIF-PT (Conformer encoder) & 89.60\% & 78.67\% \\
CIF-PT (Data2vec encoder) & \textbf{91.32\%} & \textbf{81.63\%} \\
\hline
\end{tabular}
}
\caption{\label{main_results1}
Comparison with the published results on SLU benchmark (SLURP), including two tasks: intent classification (IC) and slot filling (SF). Our CIF-PT method uses the result of M0, M1 in Table 2, respectively. Both are pre-trained and fine-tuned only on the SLURP.
}
\end{table}

\begin{table*}
\renewcommand\arraystretch{0.85}
\centering
\resizebox{0.9\textwidth}{!}{
\begin{tabular}{c|c|c|c|c}
\hline
\textbf{(Model Id.)} & \textbf{Method} & \textbf{Speech Encoder} & \textbf{Intent Classification} & \textbf{Slot Filling}  \\
\textbf{} & \textbf{} & \textbf{} & (Acc.) & (SLU-F1)  \\
\hline
M0 & CIF-PT & Conformer & 89.60\% & 78.67\%  \\
M1 & CIF-PT & Data2vec & \textbf{91.32\%} & \textbf{81.63\%}  \\
\hline 
\multicolumn{5}{l}{On the importance of CIF-PT} \\
\hline \hline
M2 & M0 w/o any PT & Conformer & 86.43\% (-3.17\%) & 72.51\% (-6.16\%)  \\
M3 & \quad + triple steps  & Conformer & 87.28\% (-2.32\%) & 74.92\% (-3.75\%) \\
M4 &+ CTC-PT & Conformer & 86.41\% (-3.19\%) & 75.87\% (-2.80\%) \\
\hline
\multicolumn{5}{l}{On the importance of language model distillation (LMD)} \\
\hline \hline
M5 & M0 w/o LMD & Conformer & 88.31\% (-1.29\%) & 77.84\% (-0.83\%) \\
M6 & M1 w/o LMD & Data2vec & 91.18\% (-0.14\%) & 81.02\% (-0.61\%) \\ 
\hline 
\multicolumn{5}{l}{On the importance of CIF alignment (all w/o language model distillation)} \\
\hline \hline
M7 & M3 w/o CIF  & Data2vec & 90.36\% (-0.96\%) & 79.29\% (-2.34\%) \\
M8 & \quad +CTC-PT  & Data2vec & 90.63\% (-0.69\%) & 80.31\% (-1.32\%) \\
\hline
\end{tabular}
}
\caption{\label{main_results2}
Ablation study on the proposed CIF-PT. For fair comparison, models in this table use the same structure of SLU decoder. 
For M7 where CIF is ablated, it directly passes the frame-level outputs of speech encoder to the SLU decoder. For M8, it follows the model structure of M7 but performs CTC Pre-Training (CTC-PT) on ASR tasks before training on SLU. The model structure of M4 is similar to M8 except using conformer as its speech encoder. All models are pre-trained and fine-tuned only on the SLURP data.
}
\end{table*}

\subsection{Training and Evaluation}
We use an AdamW \citep{loshchilov2018decoupled} optimizer with $\beta_1$ = 0.9, $\beta_2$ = 0.98 and weight decay of 1e-5. During CIF pre-training, we warm up the learning rate for the first 4\% of updates to a peak of 1e-3 and keep it constant in the later 64\% of updates, then linearly decay it to 1e-4. The number of total training steps is 80k. We set the weight of CTC loss $\lambda_1=0.5$, and the weight of quantity loss $\lambda_2=1.0$. The hyper-parameter of LMD loss is explored in section \ref{comparison_LMD}. During SLU training, we follow the Noam scheduler \citep{vaswani2017attention} with 1600 warm-up steps and peak learning rate of 5e-4. The number of total training steps is 32k. 

After training, we first perform model average on the last 10 checkpoints for all models and then use the averaged model for evaluation. We follow the metric of accuracy and SLU-F1 \citep{bastianelli2020slurp} to evaluate the models on task of IC and SF, respectively. During the inference of SF task, we perform beam search with beam width 10 and a temperature scalar of 1.25 
. All experimental results are averaged at least 2 runs.

\section{Results and Analysis}
\label{result}
\subsection{Main Results}
\label{sec_main_results}

To verify the effectiveness of our proposed methods, we first conduct three sets of experiments to explore the importance of designs in CIF-PT. The main results are summarized in Table \ref{main_results2}. 

The first two rows of Table \ref{main_results2} show the performance of our end-to-end SLU models using CIF-PT. Consistent with our expectation, the model M1 with the self-supervised data2vec encoder obtains better results than the model M0 with conformer encoder on both tasks. We also compare the performance of our methods with the published results. As shown in Table \ref{main_results1}, the model with CIF-PT (M1 in Table \ref{main_results2}) achieves state-of-the-art result on both of IC and SF tasks. The performance advantages on the task of SF reaches 2.71\% SLU-F1. We suspect that the cross-modal representation extracted by CIF-PT contains more language knowledge that benefits more to SF, which needs to predict the slot key and speech content simultaneously
. It is worthy to mention that the model M0 with conformer encoer also achieves competitive performance, which is even superior or comparable to the published strong models \citep{wang2021fine, seo2022integration} with self-supervised speech encoder.

For the model of M2 in Table \ref{main_results2}, we ablate CIF-PT utilized in the model M0 and conduct a joint training of ASR and SLU tasks from scratch. The results show that ablating CIF-PT leads to a large performance degradation on both SLU tasks. Since CIF-PT consumes extra pre-training steps, we suspect the total training step maybe a factor of the performance gap. Therefore, we increase the training step to triple (from 32k to 96k) to obtain the model M3. The performance gap is narrowed but the model M0 with CIF-PT still has a certain performance advantage over model M3 with longer SLU training.

\begin{figure*}[htbp]
\centering
\subfigure[Comparison of LMD methods on IC]{
\begin{minipage}[t]{0.33\linewidth}
\centering
\includegraphics[width=2in]{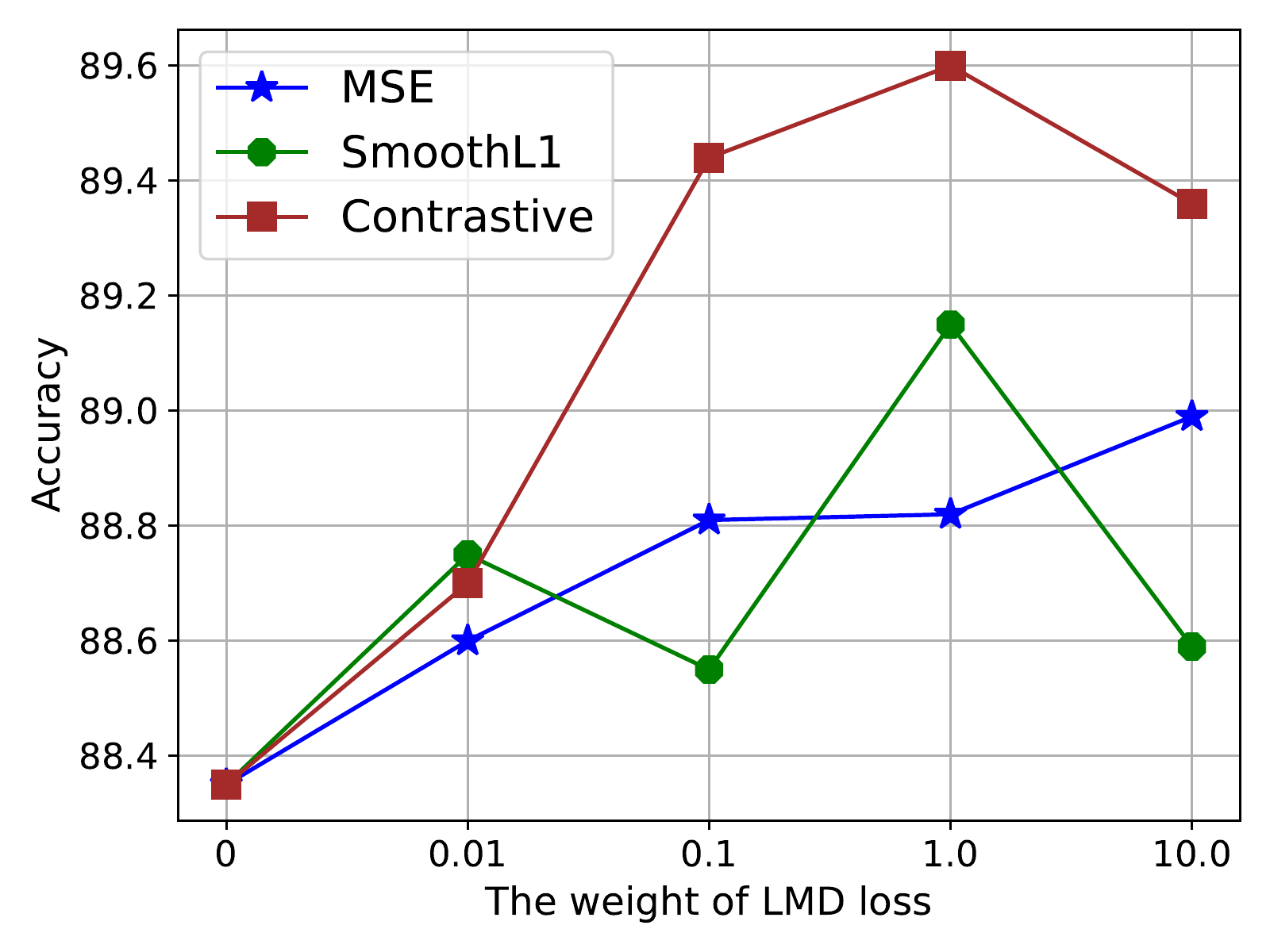}
\end{minipage}%
}%
\subfigure[Comparison of LMD methods on SF]{
\begin{minipage}[t]{0.33\linewidth}
\centering
\includegraphics[width=2in]{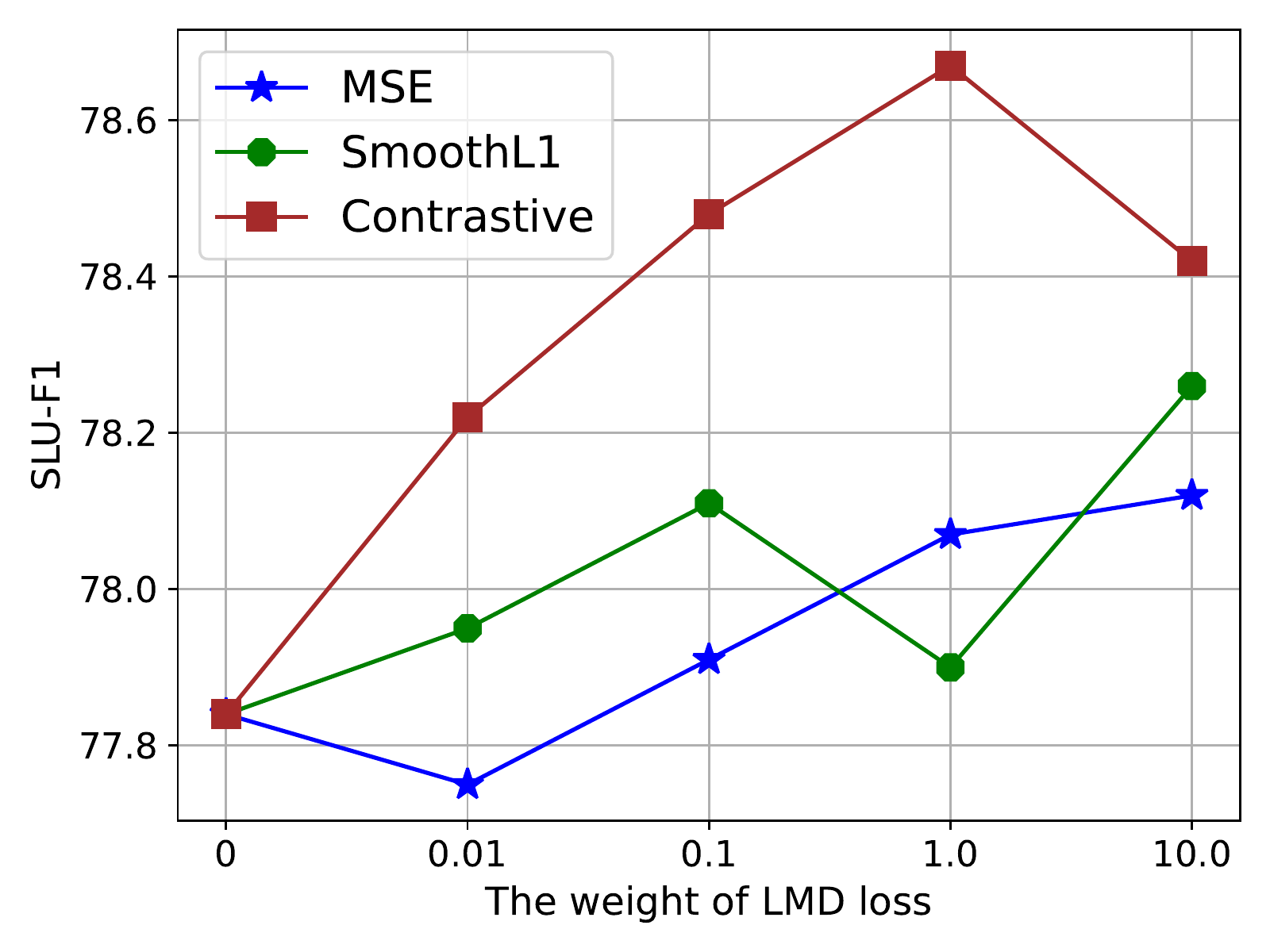}
\end{minipage}%
}%
\subfigure[Scalar $\tau$ in contrastive LMD]{
\begin{minipage}[t]{0.33\linewidth}
\centering
\includegraphics[width=2in]{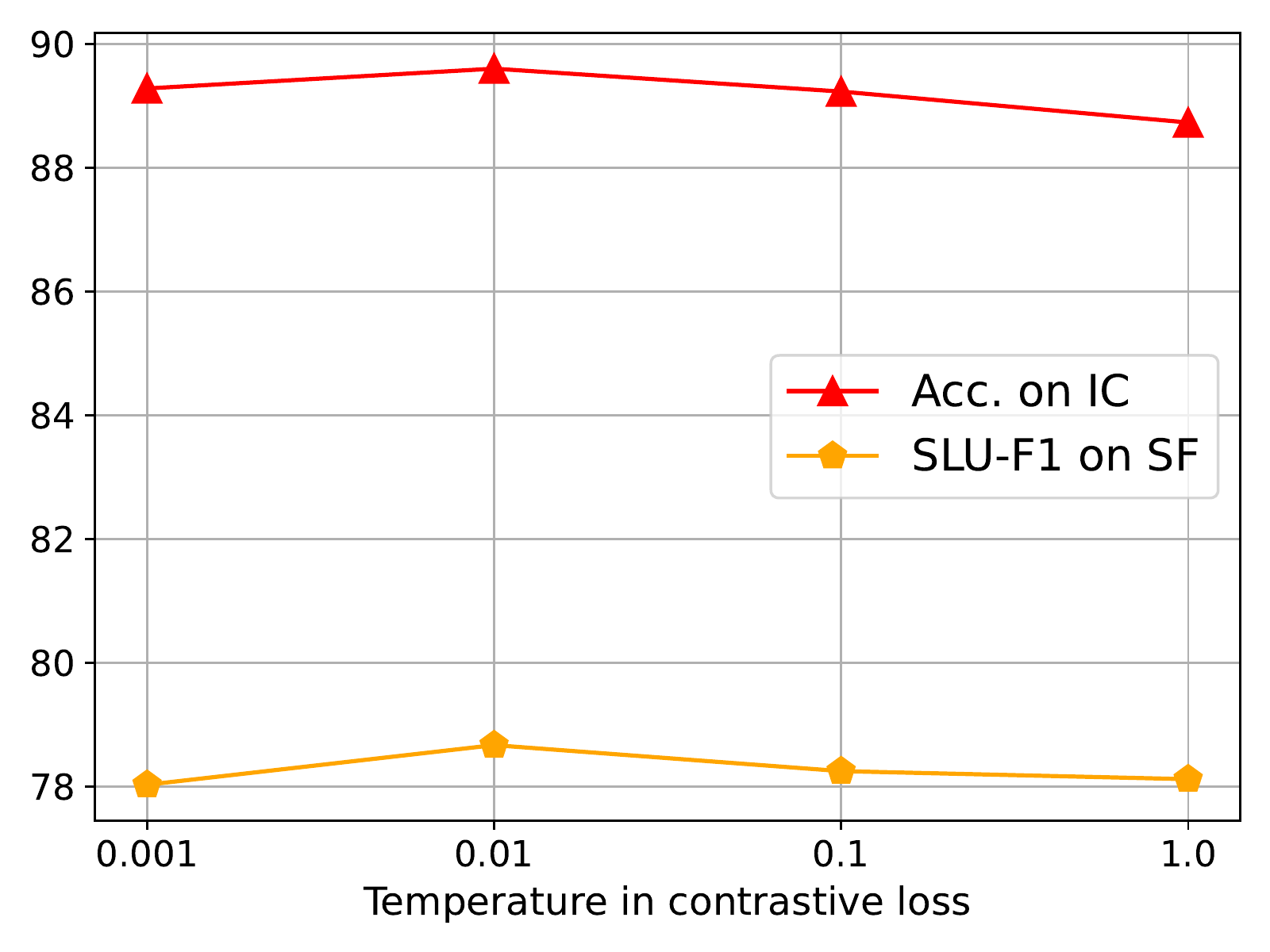}
\end{minipage}
}%
\centering
\caption{\label{LMD_figure}
(a) and (b) depict the performance fluctuation of different LMD methods on the two SLU tasks as the weight $\lambda$ of LMD loss changes. (c) depicts the performance fluctuation of the contrastive LMD method as the temperature scalar $\tau$ changes.}
\end{figure*}

For the model of M5 and M6 in Table \ref{main_results2}, we ablate language model distillation (LMD) utilized in CIF-PT. During pre-training, we find applying LMD bring  3.9\% (14.83 $\rightarrow$ 14.25) relative WER reduction on the model with conformer encoder.
During SLU training, we also observe the introduced LMD methods boosts the performance improvements on the two tasks in Table \ref{main_results2}. For the reason of the smaller performance improvements of data2vec encoder 
, it may be that the model with data2vec encoder itself has strong modeling power and already learns effective pattern and textual knowledge, so that the injected textual knowledge can only be helpful for fewer evaluation samples.

We also compare the cross-modal representation extracted by CIF-PT with the speech representation derived from self-supervised learning. For the model M7 in Table \ref{main_results2}, we ablate the frame-to-token CIF alignment in SLU models and directly pass the frame-level speech representation extracted by data2vec to the SLU decoder. 
Although achieving competitive results, model M7 could achieve further improvements after combining with CIF. We suspect the reason is two-folds: 1) CIF performs frame-to-text mapping that integrates relevant speech/semantic information, thus able to remove information redundancy in adjacent frames, 2) CIF-PT bridges the speech representation and text representation through ASR training and LMD, thus providing more textual knowledge that benefits SLU performance. To further verify our hypothesis, we introduce CTC-based ASR pre-training (CTC-PT) before the training of SLU model. Results show that CTC-PT provides improvements on SLU tasks (M7 $\rightarrow$ M8, M2 $\rightarrow$ M4), but it still has gap from CIF-PT
. Above observations demonstrate the effectiveness of CIF-PT. 

\subsection{Comparison on Language Model Distillation}
\label{comparison_LMD}

In this part, we compare different language model distillation (LMD) methods applied in CIF-PT. From the Figure \ref{LMD_figure} we get three observations: (1) All LMD methods provide positive effects on SLU performance in most cases, except for one outlier uses MSE loss with a weight of 0.01 on SF. 
The degradation disappears as the loss weight increases; (2) The contrastive LMD method shows better quality on both SLU tasks than the other two methods. 
We suspect the reason is contrastive distillation with proper temperature
scalar mainly focuses on distinguishing hard negatives, instead of forcing the representation to be consistent like MSE. This helps the extracted representation retain speech and language information at the same time, which may benefit SLU modeling; (3) Different temperature scalar in contrastive LMD method has effects on down-streaming SLU tasks, with a $\tau$ value of 0.01 producing the best results on both SLU tasks. 

\subsection{Comparison on Neural Interfaces}
\label{neural_interface}
We have compared the token-level representation $c_i$ extracted by CIF-PT with frame-level speech representations in section \ref{sec_main_results}. In this part, we continue to compare $c_i$ with other popular token-level neural interfaces (or representations) summarized in \citep{Raju2022}, including hidden interface $m_i$, posterior interface $p_i$, tied embedding interface $e_i$ and the combinations. For fair comparison, we give up using LMD loss in CIF-PT which benefits $c_i$. The results are shown in Table \ref{neural_interface}.

On the task of IC, we find all token-level neural interfaces achieve comparable accuracy. This may be because these interfaces contain close information that is useful for IC, and the pooling operation in intent decoder further reduce the discrimination between representations. The combination of $c_i$ and $e_i$ achieves the best performance. We suspect this is because they are located at the beginning ($c_i$) and end ($e_i$) of ASR decoder respectively, so they may have a large information difference and complementarity. 

\begin{table}
\centering
\begin{tabular}{c|c|c|c}
\hline
\multirow{2}*{\textbf{Interfaces}} & \multirow{2}*{\textbf{IC } (Acc.)} & \multicolumn{2}{c}{\textbf{SF} (SLU-F1)}   \\
\cline{3-4}
 &  & generation & tag \\
\hline
$c_i$ & 88.31\%  & \textbf{77.84\%} &  71.68\% \\
\hline \hline
$m_i$ & 88.26\%  & 68.24\%  & 74.21\% \\
$e_i$ & 88.25\%  & 65.42\%  & 74.42\% \\
$p_i$ & 88.30\%  & 54.04\% & 72.94\% \\
$c_i,m_i$ & 88.48\%  & 74.94\% & \textbf{74.61\%} \\
$c_i,e_i$ & \textbf{88.79\%}  & 76.22\% & 74.52\% \\
\hline
\end{tabular}
\caption{\label{neural_interface}
Comparison with other token-level neural interfaces summarized in \citep{Raju2022}. Here, $c_i$ represents the cross-modal representations extracted by CIF-PT. $m_i$ represents the output of ASR decoder (Hidden Interfaces). $p_i$ represents the posterior predicted by ASR decoder (Posterior Interface). $e_i$ represent the token embedding of ASR's one-best token sequence (Tied Embedding Interface). Generation and tag in the table describe two types of slot decoder used in our model, we detail their structures in Section \ref{subsec:details}.
}
\end{table}

On the task of SF, we observe a relatively large differentiation among these neural interfaces. On the model using generation-based slot decoder, $c_i$ obtains the best performance, while other interfaces have a certain performance gap in comparison. This phenomenon can be understood as $c_i$, which is sourced from pure speech inputs, contains more original and comprehensive speech information. It can provide sufficient information for the calculation of cross-attention in the slot decoder. In contrast, the other interfaces are all calculated via the autoregressive ASR decoder, thus the information may be biased to a certain hypothesis with errors in inference. In addition, using $c_i$ as the interface can also avoid the mismatch between the teacher-forcing inputs and predicted inputs in inference.

Interestingly, on the model using tag-based slot decoder, $c_i$ performs inferior to other neural interfaces. Since the tag-based slot model predicts slot key for each token of the one-best ASR hypothesis, the neural interfaces $m_i$, $p_i$, $e_i$ that are updated synchronously with the ASR decoding could provide closer slot prediction for the final ASR hypothesis. The original speech information provided by $c_i$ can also provide supplements to these interfaces, and the best performance is obtained by the combination of $c_i$ and $m_i$.

Between two types of slot decoder, the model with generation-based slot decoder is superior to the tag-based slot decoder, we believe this is because generation-based decoder utilize the bi-directional contextual information from full sequence, which makes it have higher ceiling in the prediction of slot information. In contrast, tag-based decoder could only use the uni-directional information that is limited by the autoregressive ASR decoder. However, this characteristic makes the tag-based model suitable for the application scenario with low-latency. 

\subsection{Comparison on Out-of-domain Data}
\label{ood_data}
In above experiments, CIF pre-training is performed on the in-domain SLURP dataset. During SLU training, the pre-trained parameters are kept frozen (`Slurp-Frozen' in Table \ref{cifpt_set}) and only the part of SLU decoder is trained. In this part, we first explore unfreezing the pre-trained parameters during SLU training (`Slurp-Unfrozen' in Table \ref{cifpt_set}). Specifically, we hold the pre-trained parameters frozen in the first half of training, and then make the model entirely trained by performing joint training of ASR and SLU tasks. Results show that unfreezing pre-trained parameters leads to slight performance degradation. We suspect it is because the textual knowledge injected by LMD suffers catastrophic forgetting during SLU training. But the result achieved by `Slurp-Unfrozen' is still better than the model using the frozen pre-trained model without LMD (model M5 in Table \ref{main_results2}).

\begin{table}
\centering
\begin{tabular}{l|c|c}
\hline
Model & \textbf{IC } (Acc.) & \textbf{SF} (SLU-F1)  \\
\hline
Slurp-Frozen & 89.60\% & 78.67\% \\
Slurp-Unfrozen & 88.84\% & 78.08\%  \\
LS-Frozen & 80.65\% & 64.02\% \\
LS-Unfrozen & \textbf{90.65\%} & \textbf{79.74\%} \\
\hline
\end{tabular}
\caption{\label{cifpt_set}
Comparison on the out-of-domain data. In the column of model, Slurp and LS before the dash represent the utilized pre-trained dataset, LS represents Librispeech. Frozen and Unfrozen represent the state of pre-trained parameters in SLU training.
}
\end{table}

We also conduct experiments on an out-of-domain pre-training dataset (Librispeech) to explore its effects on the final SLU performance. Consistent with our expectations, freezing the parameters pre-trained on out-of-domain data (`LS-Frozen' in Table \ref{cifpt_set}) leads to a large performance degradation on SLU tasks of SLURP. When unfreezing these pre-trained parameters (`LS-Unfrozen' in Table \ref{cifpt_set}) during SLU fine-tuning, the model obtains a noticable performance boost, and even outperforms the model achieved on slurp dataset. This partly reflects the good generalization and the potential on transfer learning of our proposed CIF-PT method.

\section{Conclusion}
In this work, we propose a new pre-training paradigm: Continuous Integrate-and-Fire Pre-Training (CIF-PT) for end-to-end SLU. CIF serves as a bridge connecting speech and text modality: on the one hand, it integrates speech representation into token-level through its frame-to-token alignment ability learned from ASR pre-training task. On the other hand, it support one-to-one transfer of the textual knowledge into the integrated token-level speech representation via the pre-training of language model distillation. After CIF-PT, we obtain a cross-model representation that is used as neural interface into down-streaming SLU tasks. 

Evaluated on the largest SLU benchmark of SLURP, CIF-PT creates new state-of-the-art result on both of IC and SF tasks. We further validate the effectiveness and generalization of CIF-PT by a series of experiments including ablation study and the pre-training on out-of-domain data. We also observe the cross-modal representation extracted by CIF-PT shows its competitiveness in comparison with other neural interfaces on SLU. We believe that CIF-PT has the potential to better encode long-form speech content (e.g. spoken paragraph) through its language model distillation, and will explore to combine it with LLM methods like ChatGPT to further empower spoken language understanding (SLU) systems.

\section{Limitation}
In the process of conducting experiments, we find our method has some limitations. First, CIF-PT needs to be performed on the dataset with speech-text pair. For some small-scale dataset that only contains speech and SLU labels, our method needs to use external ASR dataset to conduct the pre-training, leading to the increase of complexity of model building. In addition, in CIF-PT, we need to ensure that the tokenizer of the pre-trained language model is consistent with the tokenizer in the ASR task. However, there is usually a gap between the two in terms of vocabulary size. In consideration of performance, it is necessary to modify the tokenzier of one or both sides.

\bibliography{anthology,custom}
\bibliographystyle{acl_natbib}



\clearpage
\appendix
\section{Appendix}
\label{sec:appendix}

\subsection{Computational Experiments}
The total parameters for our SLU model with conformer encoder is 95.07 M. It costs 10.1 hours and 12.0 hours for CIF-PT and SLU fine-tunig on 8 A100 GPUs, respectively. The batch size of both stages is set to 30000 frames on each GPU. For our CIF SLU model with data2vec encoder, it has 357.50M parameters and needs 23.0 hours and 7.5 hours to finish CIF-PT and SLU fine-tuning, the corresponding batch size for the two stages is set to 1.2M and 1.6M samples, respectively.

\subsection{Details of Model Structure}
\label{subsec:details}
\begin{figure}[htbp] 
	\centering 
	\includegraphics[width=\textwidth]{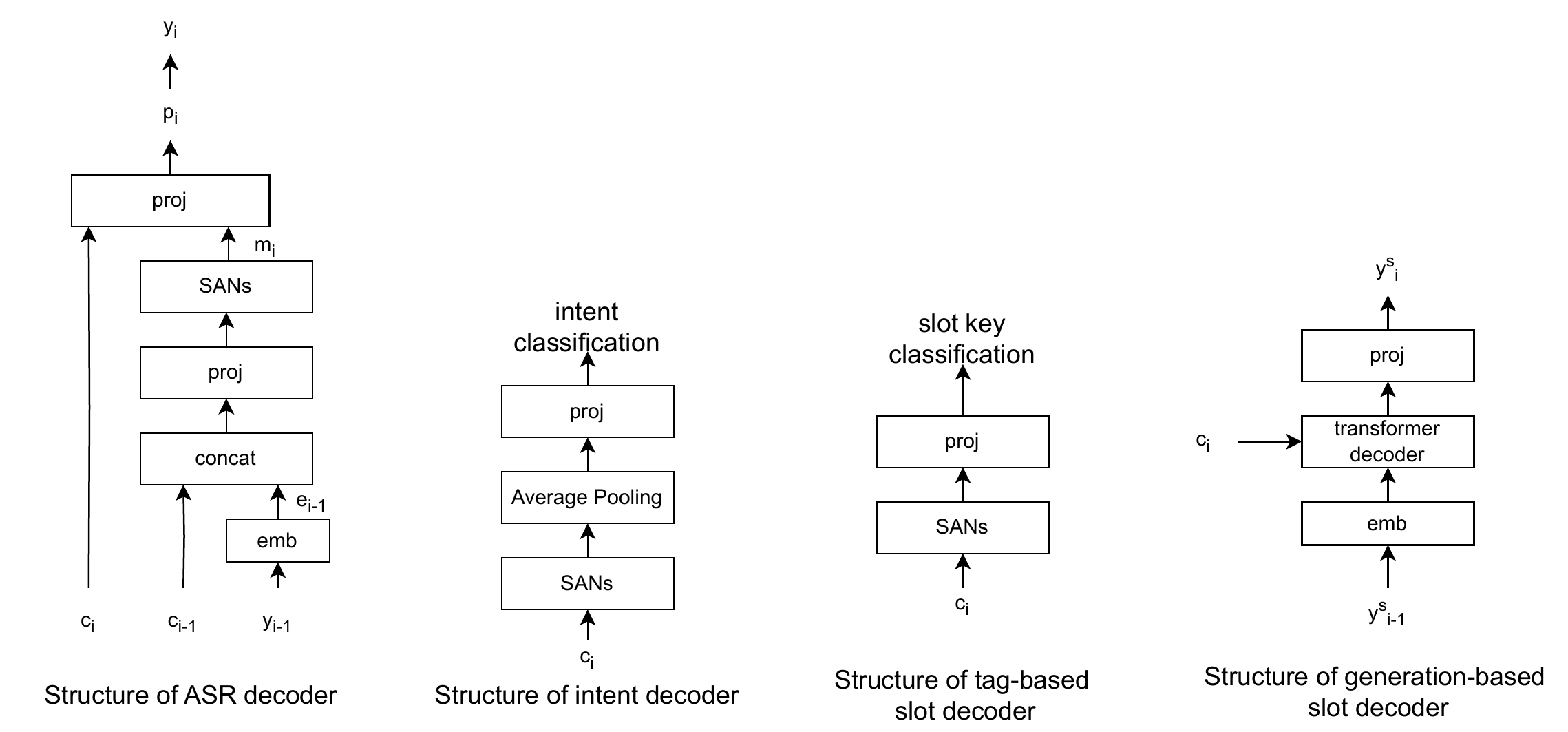}
	\caption[width=\textwidth]{Model structure of our ASR decoder and SLU decoders. Different neural interfaces are depicted in the ASR decoder. The details of tag-based slot decoder and generation decoder are also included in this figure.  $c_i$ in this figure could be replaced by other interfaces, which are investigated in Table \ref{neural_interface}.}.
	\label{detail111} 
\end{figure}


\end{document}